\documentclass[conference]{IEEEtran}
\IEEEoverridecommandlockouts
\usepackage{cite}
\usepackage{amsmath,amssymb,amsfonts}
\usepackage{algorithmic}
\usepackage{graphicx}
\usepackage{textcomp}
\usepackage{xcolor}
\def\BibTeX{{\rm B\kern-.05em{\sc i\kern-.025em b}\kern-.08em
    T\kern-.1667em\lower.7ex\hbox{E}\kern-.125emX}}

\usepackage{url}
\usepackage{multirow}

\begin{document}

\title{Scaling Multimodal Search and Recommendation with Small Language Models via Upside-Down Reinforcement Learning}

\author{\IEEEauthorblockN{Yu-Chen Lin}
\IEEEauthorblockA{Adobe}
\and
\IEEEauthorblockN{Sanat Sharma}
\IEEEauthorblockA{Meta}
\and
\IEEEauthorblockN{Hari Manikandan}
\IEEEauthorblockA{Adobe}
\and
\IEEEauthorblockN{Jayant Kumar}
\IEEEauthorblockA{Adobe}
\and
\IEEEauthorblockN{Tracy Holloway King}
\IEEEauthorblockA{Adobe}
\and
\IEEEauthorblockN{Jing Zheng}
\IEEEauthorblockA{Adobe}
}

\maketitle

\begin{abstract}
In this work, we investigate how small language models (SLMs) can be scaled to support multimodal search and recommendation use cases while remaining efficient enough for real-time, resource-constrained deployments. We present a framework that combines upside-down reinforcement learning with synthetic data distillation from a large language model (Llama-3 \cite{llama3}) to train a 100M-parameter GPT-2 model \cite{gpt2} for multitask prompt generation. Despite being up to 80 times smaller than state-of-the-art large language models (LLMs), our SLM achieves relevance and diversity scores within 6\% of competitive baselines such as Llama-3 8B, Qwen3 8B, and Ministral 8B. These results demonstrate that SLMs can effectively handle multimodal search and recommendation tasks, while dramatically reducing inference latency and memory overhead. Our study highlights the potential of lightweight models as practical engines for scalable multimodal discovery, bridging the gap between cutting-edge research and real-world multimodal applications such as media recommendations and creative content generation.
\end{abstract}


\begin{figure*}[!tb]
\centerline{\includegraphics[width=\textwidth]{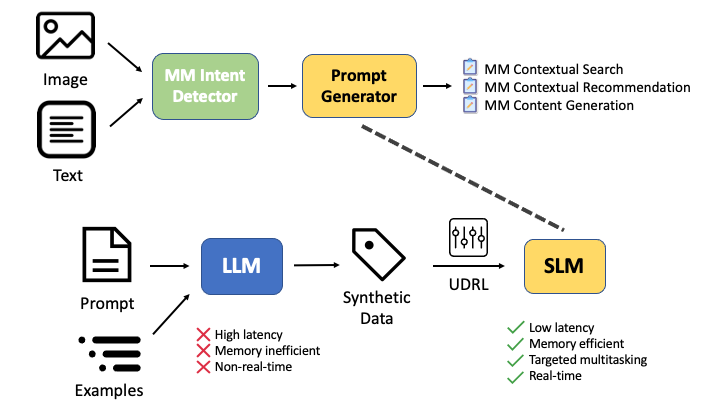}}
\caption{The top workstream illustrates our multimodal system with intent detection and prompt generation for contextual search, recommendation, and content generation. The bottom workstream shows our pipeline for training a small language model (SLM) via synthetic data distillation and upside-down reinforcement learning, enabling an efficient real-time alternative to large language models.}
\label{fig:framework}
\end{figure*}

\section{Introduction}

Multimodal search and recommendation has emerged as a central capability for applications such as e-commerce, media platforms, and interactive agents. With the revolution of large language models (LLMs), more powerful multimodal use cases have become possible, yet these gains come with substantial computational and memory costs, especially at inference time. These limitations hinder their deployment in high-frequency, real-time applications where efficiency is critical. To address this challenge, we propose a small language model (SLM) framework that serves as an efficient and effective multitask learner for multimodal prompt generation, leveraging upside-down reinforcement learning (UDRL) to control generation.

Fig.~\ref{fig:framework} provides an overview of our approach. Our multimodal system accepts both image and text inputs, which are processed by an in-house detector for intent understanding before passing to a prompt generator. The generated prompts, in turn, enable multiple downstream use cases. For example, they can power contextual search, content generation suggestions, and multimodal recommendations by producing relevant text or image outputs aligned with user intent. To train the prompt generator, we employ a compact SLM. Synthetic data is first generated from a large language model (Llama-3) using prompt engineering and few-shot examples, and then distilled into the SLM via UDRL \cite{udrl0, udrl}. This produces a specialized model optimized for low latency, memory efficiency, targeted multitasking, and real-time multimodal generation.

Our approach is built on three core contributions:

\begin{itemize}
    \item SLMs as Efficient Multitask Learners: We demonstrate that SLMs, trained with targeted synthetic data, can support diverse multimodal discovery tasks while bridging the gap between LLMs and deployable engines.
    \item A scalable training pipeline with UDRL: We introduce a novel pipeline that distills supervision from LLMs and applies UDRL to optimize SLMs for prompt generation across multiple tasks, achieving competitive relevance and diversity.
    \item Synthetic Dataset Distillation: Leveraging LLM-generated data, we curate high-quality training examples that enable SLMs to learn effectively from minimal resources.
\end{itemize}

Through systematic experiments, we show that our SLM can achieve performance comparable to much larger LLMs with an 80-fold reduction in model size and delivering substantial gains in latency and memory efficiency. This makes SLMs suitable for real-time multimodal discovery applications, where responsiveness and scalability are essential. Moreover, our framework is highly adaptable and can be integrated with commercial text-to-image generation systems (e.g., Adobe Firefly), enabling practical deployment in diverse real-world scenarios.

\section{Methodology}

As illustrated in Fig.~\ref{fig:framework}, our task involves generating prompts for generative models based on multimodal inputs. The process begins with detecting user intents using an in-house intent detector. Once the intents are identified, we aim to generate prompts accordingly. Rather than relying solely on a large language model (LLM) or a multimodal LLM, we adopt a more efficient approach.

Specifically, our methodology focuses on training a small language model (SLM) for prompt generation with intents that emphasizes high efficiency and multitask capabilities. We achieve this by leveraging synthetic data distillation from a large language model (LLM) combined with upside-down reinforcement learning.

\subsection{Synthetic Dataset Distillation}

To enable the SLM to capture complex task knowledge from a larger model, we first create a synthetic dataset using Llama-3. This involves generating high-quality, task-specific data that allows the SLM to learn from the representations of LLM effectively. We utilize vLLM \cite{vllm} to parallelize our dataset generation process.

\subsubsection{Dataset Curation Process}

\begin{itemize}
    \item \textbf{Intent and Prompt Pair Generation:} We curated a dataset of 52 million intent-prompt pairs by prompting the LLM with various intent descriptions and collecting its responses. Each sample consists of an intent description paired with a generated prompt, tailored to a range of real-world tasks. For example, intents such as ``birthday celebration,'' ``holiday sale,'' or ``product launch'' were used to generate prompts that align with these contexts.

    \item \textbf{Structured Prompting for Targeted Tasks:} To ensure diversity and relevance in generated prompts, we used structured prompting with few-shot examples and task specifications. Each prompt to Llama-3 included high-level task descriptions and few-shot examples to encourage contextually appropriate and diverse outputs. Different prompting strategies are used to address the specific requirements of different tasks. By specifying factors like tone, style, and context, we created a dataset that is comprehensive and task-aligned.

    \item \textbf{Labeling and Intent Selection:} The intent descriptions were sourced from in-house creative knowledge graph and augmented with diverse asset metadata from internal images and templates. This process provided a broad set of intents that accurately represent real-world use cases. Each intent was paired with multiple generated prompts, thus ensuring a rich dataset of prompt variations for each concept.
\end{itemize}

\begin{table*}[!tb]
\caption{Examples of intents and generated prompts for image and template generation, including format specifications. Having task, length, and intent tags helps guide the generations.}
\begin{center}
\begin{tabular}{|p{3cm}|p{3.5cm}|p{8.5cm}|}
\hline
\textbf{Intents / Task} & \textbf{Generated Prompt} & \textbf{Training Data Format} \\ \hline
Topic: birthday \newline Scene object: balloon \newline Task: prompt for text-to-image generation & Whimsical birthday celebration featuring giant balloons in fun shapes and sizes, tied to a birthday child's arm or wrist. & \texttt{<|19|> <|intent|> Topic: birthday, Scene object: balloon <|IP|> whimsical birthday celebration featuring giant balloons in fun shapes and sizes, tied to a birthday child's arm or wrist.} \\ \hline
Topic: birthday party \newline Design type: invitation \newline Task: prompt for text-to-template generation & Create a whimsical birthday party invitation template with balloons, confetti, and a playful theme. & \texttt{<|14|> <|intent|> Topic: birthday party, Design Type: invitation <|TP|> create a whimsical birthday party invitation template with balloons, confetti, and a playful theme.} \\ \hline
\end{tabular}
\label{table:intents_prompts}
\end{center}
\end{table*}

This curated dataset of intent-prompt pairs serves as a distilled form of knowledge from Llama-3, allowing the SLM to learn diverse language patterns and contextual nuances without directly training on a massive language model. Examples of intents to generated prompts are shown in Table~\ref{table:intents_prompts}.

\subsection{Upside-Down Reinforcement Learning for SLM Optimization}\label{ssec:udrl}

To optimize the SLM's generation quality and control specific attributes like length and relevance, we utilize upside-down reinforcement learning (UDRL) \cite{udrl0, udrl}. This approach allows the model to learn specific objectives based on desired outcomes rather than traditional reward structures.

\subsubsection{Reward-Based Prompt Generation}

Upside-down reinforcement learning frames the prompt generation task as an optimization problem where the SLM aims to achieve target specifications for each generated output. This process is as follows:

\begin{itemize}
    \item \textbf{Controlled-Length Generation:} The SLM is trained to produce prompts within a desired length range (e.g., 10 to 35 words). Tokens indicating target lengths are incorporated into the input, guiding the model towards generating responses that match the specified length with a mean squared error consistently under two words. More evaluations are in Section~\ref{sec:eval}.

    \item \textbf{Modality-Agnostic Prompting:} We trained the SLM to handle both text-to-image (T2I) and text-to-template (T2T) prompts within the same framework. This was achieved by adding modality tokens to each training instance, allowing the model to distinguish between generation tasks and tailor its output accordingly.

    \item \textbf{Contextual Relevance and Specificity:} By assigning relevance scores to generated prompts based on a predefined metric (e.g., alignment with target intent or clarity), the model learns to prioritize contextually relevant and specific responses. During training, prompts that meet these objectives are positively reinforced, improving the SLM's ability to generate accurate and contextually appropriate prompts. While we do not implement this aspect in our current pipeline, it could be valuable for other applications.
\end{itemize}

\subsection{Model Architecture and Key Capabilities}

Our SLM is based on nanoGPT\footnote{\url{https://github.com/karpathy/nanoGPT}}, a compact variant of GPT-2, with 104 million parameters, configured with 12 layers, 12 attention heads, and an embedding dimension of 768. The model architecture and training setup are designed to maximize efficiency and multitask performance.

\subsubsection{Model Specifications}\label{sssec:spec}

\begin{itemize}
    \item \textbf{Parameter Efficiency:} The SLM contains approximately 1/80th the parameters of Llama-3 8b and similar state-of-the-art LLMs, making it computationally efficient and suitable for deployment on standard hardware. 
    
    \item \textbf{Inference Speed:} Our SLM achieves a processing speed of up to 338 tokens per second on a single A10G GPU (non-batched, non-quantized, or accelerated), making it suitable for real-time applications. This performance is especially advantageous in resource-constrained environments where inference latency is a critical factor.

    \item \textbf{Multitask Learning Capabilities:} The SLM is trained to handle both T2I and T2T prompts, making it a versatile tool for multimodal applications. Through the integration of task-specific tokens, the model can generate contextually accurate prompts tailored to the input task, whether it’s for text-to-image generation or template-based design.
\end{itemize}

\begin{table*}[!tb]
\caption{Relevance scores for prompt generation tasks, evaluated on text-to-image and text-to-template prompts under both few-shot and zero-shot prompting. Our SLM model, which operates in a zero-shot setting, demonstrates competitive performance with much larger models. With the UDRL training paradigm, SLM achieves efficiency suitable for enterprise use, outperforming or matching models with significantly higher parameter counts.}
\begin{center}
\begin{tabular}{|l|c|c|c|c|c|c|c|}
\hline
\multirow{2}{*}{\textbf{Model}} & \multicolumn{2}{c|}{\textbf{Few-Shot}} & \multicolumn{2}{c|}{\textbf{Zero-Shot}} & \multicolumn{2}{c|}{\textbf{\# Tokens Per Second}} & \multirow{2}{*}{\textbf{\# Params}} \\ \cline{2-7}
& \textbf{text-to-image} & \textbf{text-to-template} & \textbf{text-to-image} & \textbf{text-to-template} & \textbf{Rate} & \textbf{Memory Used} & \\ \hline
\textbf{SLM (Ours)} & N/A & N/A & 8.31 & 8.45 & \textbf{353} & \textbf{$\sim$500MB} & \textbf{104M} \\ \hline \hline
\textbf{Llama-3 8B \cite{llama3}} & 8.63 & 8.56 & 8.36 & 8.33 & 80 & \multirow{7}{*}{\shortstack{$\sim$80GB \\ (with vLLM)}} & 8.0B \\ \cline{1-6} \cline{8-8}
\textbf{Ministral 8B \cite{mistral}} & 8.55 & 8.40 & 7.87 & 7.18 & 78 & & 8.0B \\ \cline{1-6} \cline{8-8}
\textbf{Gemma-3 4B \cite{gemma3}} & 8.66 & 8.41 & 8.51 & \textbf{8.49} & 100 & & 4.3B \\ \cline{1-6} \cline{8-8}
\textbf{SmolLM3 3B \cite{smollm3}} & 8.70 & \textbf{8.68} & 8.19 & 8.35 & 121 & & 3.1B \\ \cline{1-6} \cline{8-8}
\textbf{Llama-3.2 3B \cite{llama3}} & 8.65 & 8.60 & 8.40 & 8.44 & 142 & & 3.2B \\ \cline{1-6} \cline{8-8}
\textbf{Phi-4-mini \cite{phi4}} & 8.59 & 8.45 & 8.27 & 8.15 & 118 & & 3.8B \\ \cline{1-6} \cline{8-8}
\textbf{Qwen3 8B \cite{qwen3}} & \textbf{8.80} & 8.63 & \textbf{8.52} & 8.37 & 76 & & 8.2B \\ \hline
\end{tabular}
\label{table:relevance_scores}
\end{center}
\end{table*}

\subsubsection{Training with UDRL}

The training data are formatted as \texttt{<|\# words of the prompt|> <|intent|> INTENT <|prompt for T2I (IP) or T2T (TP)|> PROMPT}. As outlined in Section~\ref{ssec:udrl}, we combine the word count and modality tokens to create a single training instance; refer to Table~\ref{table:intents_prompts} for examples.

We utilize a vocabulary set with legal approval, ensuring it excludes any offensive, discriminatory, or biased language. We then train a BPE tokenizer with 25,600 tokens from scratch using our curated dataset. Subsequently, we train the nanoGPT model using next-token prediction, employing four A10G GPUs over 10 days, completing approximately 300,000 iterations with a batch size of 128 and a learning rate of 6e-4.

By combining this training approach with upside-down reinforcement learning, we ensure that our SLM can effectively manage length control and multitasking. During inference, we can easily control output length by specifying a length token and define the desired task using the modality token. Please refer to Section~\ref{sec:eval} for more evaluations of these abilities.

\section{Experiments and Evaluations}\label{sec:eval}

We focus on both qualitative and quantitative evaluations to judge the quality of our model.

\subsection{Quantitative Evaluation}

The quantitative evaluation is divided into two primary areas: relevance evaluation, which assesses how well the generated prompts align with the specified queries and modality requirements, and task adherence evaluation, which measures the SLM’s accuracy in meeting specific prompt length requirements.

\subsubsection{Relevance Evaluation}

To evaluate relevance, we conducted experiments to measure how accurately the SLM-generated prompts aligned with the specified queries and task requirements. Relevance was assessed using an automatic method with LLM-as-a-judge, where GPT-4o-mini \cite{gpt4} served as the evaluator. Each generated prompt was rated on a scale from 1 to 10, where higher scores indicate stronger alignment with the target query. We provided several examples and criterion to the LLM judge prior to the evaluation. Some of these include 
\begin{itemize}
    \item Correctness - does the prompt contain grammatical or semantic errors.
    \item Clarity - is the prompt clear to understand, is the grammar structure sound.
    \item Completeness - does the prompt utilize all of the provided query context.
    \item Usefulness - is the prompt generated useful for the task provided.
\end{itemize}

For comparison, we evaluated our SLM with state-of-the-art LLMs ranging in size from 3B to 8B. Table~\ref{table:relevance_scores} displays the relevance scores for both text-to-image (T2I) and text-to-template (T2T) prompt generation tasks. We investigated the performance of these LLMs under both few-shot and zero-shot prompting. For our trained SLM, there is no need for demonstration, which places it in the zero-shot setting. We observe the following.
\begin{enumerate}
    \item \textbf{Effective Distillation:} In the few-shot setting, our SLM achieves relevance scores nearly on par with the teacher model, Llama-3 8B, with only a minor performance gap of approximately 3\%. In the zero-shot setting, the performance of our SLM matches Llama-3 8B, and it even outperforms Llama-3 8B in T2T prompt generation. This illustrates the success of our distillation process, effectively transferring high-performance prompt generation capabilities to a more efficient model, without the need for task-specific demonstrations.
    \item \textbf{Competitive Performance with Minimal Parameters:} Despite being over an order of magnitude smaller, our SLM competes closely with models between 3B–8B parameters. On T2I zero-shot, SLM scores 8.31, only slightly behind the strongest model (Qwen-3 8B, 8.52), while in T2T zero-shot it ranks second overall (8.45). Importantly, these results are achieved with just 104M parameters—two orders of magnitude fewer than most baselines—demonstrating a remarkable balance of compactness and accuracy.
    \item \textbf{Efficiency and Deployment Readiness:} SLM is highly efficient in both speed and memory. Running on a single A100 80GB, it requires only $\sim$500MB of memory while generating at 353 tokens/second. In contrast, all baseline LLMs evaluated under the vLLM runtime require $\sim$80GB of GPU memory and achieve significantly slower generation rates (76–142 tokens/second). This means SLM is nearly 3–5$\times$ faster while being vastly more memory-efficient. Even compared to the smallest baseline (Llama-3.2 3B), SLM delivers superior or comparable prompt relevance with a model size that is over 30$\times$ smaller.
\end{enumerate}

In summary, our SLM delivers contextually accurate prompt generation while being substantially more efficient and lightweight than existing LLMs. Its strong performance across both T2I and T2T tasks, combined with extreme efficiency in speed and memory, makes it a compelling solution for real-world deployment, especially in resource-constrained environments for multimodal discovery applications.

\begin{figure*}[!tb]
\centerline{\includegraphics[width=\textwidth]{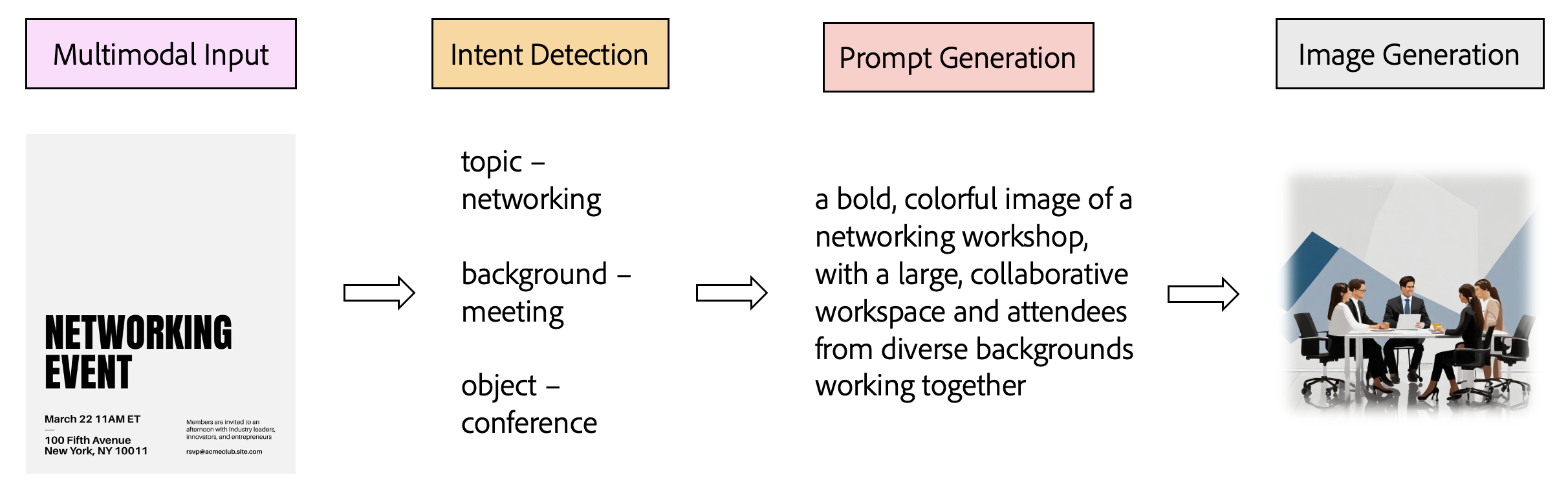}}
\caption{A sample pipeline for multimodal contextual recommendation incorporating our SLM model for prompt generation.}
\label{fig:contextual}
\end{figure*}

\subsubsection{Task Adherence Evaluation (Length)}

\begin{table}[!tb]
\caption{Task adherence results for specified prompt lengths. MSE measures the deviation from the target length, and the final row shows the percentage of prompts within ±2 words of the target length.}
\begin{center}
\begin{tabular}{|c|c|c|c|c|}
\hline
\multirow{2}{*}{\textbf{Target Length (words)}} & \multirow{2}{*}{10} & \multirow{2}{*}{20} & \multirow{2}{*}{30} & \multirow{2}{*}{35} \\
& & & & \\ \hline
\multirow{2}{*}{\textbf{Mean Length (words)}} & \multirow{2}{*}{10.3} & \multirow{2}{*}{19.2} & \multirow{2}{*}{31.1} & \multirow{2}{*}{33.8} \\
& & & & \\ \hline
\multirow{2}{*}{\textbf{Mean Square Error}} & \multirow{2}{*}{0.365} & \multirow{2}{*}{0.881} & \multirow{2}{*}{1.131} & \multirow{2}{*}{1.179} \\
& & & & \\ \hline
\multirow{2}{*}{\textbf{\% Prompts Within ±2 Words}} & \multirow{2}{*}{98\%} & \multirow{2}{*}{96\%} & \multirow{2}{*}{93\%} & \multirow{2}{*}{95\%} \\
& & & & \\ \hline
\end{tabular}
\label{table:length_adherence}
\end{center}
\end{table}

In addition to relevance, we evaluated the SLM's ability to meet target length requirements for generated prompts. The model was trained to produce prompts within specified lengths, typically in the range of 10 to 35 words. Length adherence was measured by calculating the mean squared error (MSE) between the generated prompt length and the target length, and by recording the percentage of prompts that fell within an acceptable range (±2 words from the target length). The evaluation results are in Table~\ref{table:length_adherence}.

The SLM achieves precise control over prompt length, with MSE consistently around 1. These results indicate that the model can reliably generate prompts of desired lengths, which is essential for applications requiring specific or concise outputs. This ability provides significant flexibility across diverse multimodal use cases.

\subsection{Qualitative Evaluation}

To complement the quantitative results, we conducted qualitative evaluations with human reviewers, tailored to real-world multimodal use cases.

\subsubsection{Prompt Quality}

We chose 18 human reviewers with experience in writing prompts for image and template generative models. The reviewers assessed the SLM-generated prompts on two criteria:
\begin{itemize}
    \item Relevance - how closely does the prompt align with the query.
    \item Correctness - is the prompt clear and easy to understand, with correct grammar and sound structure.
\end{itemize}
Each criterion was rated on a scale from 1 to 3 (corresponding to not, somewhat, very, e.g., not relevant, somewhat relevant and very relevant respectively), and reviewers provided additional qualitative feedback on the prompts.

In total, we collected 180 human evaluations. The human reviewers scored the prompts with an average relevance rating of \textbf{87\%}, highlighting the model's ability to generate prompts that align well with the provided queries and the task specification. For accuracy and correctness, the SLM received a score of \textbf{96\%} somewhat or very correct, indicating that the prompts were contextually accurate, grammatically correct and generally maintained coherence to the topic. We emphasize again that the prompt, which incorporates core user intents, can be leveraged to support a wide range of multimodal tasks, including contextual search, prompt suggestion, generative autocomplete, and creative content generation.

\subsubsection{Multimodal Contextual Recommendation}

We evaluated the end-to-end experience for multimodal contextual recommendation, as illustrated in Fig.~\ref{fig:contextual}. Suppose a user is editing a canvas containing some pre-specified text; in this case, we can incorporate those multimodal inputs for intent understanding and prompt generation. The generated prompt can then be used to produce images that the user can integrate into the canvas.

For this evaluation, we recruited 17 evaluators with prior experience in editing templates. We collected 210 evaluations, asking the evaluators to rank the work-in-progress canvas alongside the generated elements. The results indicate that \textbf{86\%} of evaluators were willing to incorporate the generated images into the specified canvas, demonstrating strong user acceptance.

Overall, the qualitative feedback aligns with our quantitative findings, showing that the SLM not only achieves high relevance and adherence to content length but also produces contextually rich prompts. This analysis highlights the practical applicability of SLMs for multimodal tasks, especially in scenarios where low latency and computational efficiency are essential.

\section{Conclusion}

This work demonstrates the potential of small language models (SLMs) as efficient and versatile multitask learners. By integrating upside-down reinforcement learning with supervised training, we show that an SLM can achieve performance competitive with much larger models, such as Llama-3, while reducing model size by up to 80-fold for targeted tasks. Our evaluations indicate that SLMs can preserve high contextual relevance, precision, task adherence, and precise length control, even in resource-constrained environments.

The substantial gains in efficiency and low-latency inference highlight the practicality of SLMs for real-world deployment, particularly in enterprise scenarios where computational resources and response speed are critical. These findings underscore the promise of lightweight models as scalable engines for multimodal search, recommendation, and creative content generation.

\newpage

\bibliographystyle{IEEEtran}
\bibliography{references}

@article{gpt2,
  title={Language Models are Unsupervised Multitask Learners},
  author={Radford, Alec and Wu, Jeff and Child, Rewon and Luan, David and Amodei, Dario and Sutskever, Ilya},
  year={2019}
}

@inproceedings{vllm,
  title={Efficient Memory Management for Large Language Model Serving with PagedAttention},
  author={Woosuk Kwon and Zhuohan Li and Siyuan Zhuang and Ying Sheng and Lianmin Zheng and Cody Hao Yu and Joseph E. Gonzalez and Hao Zhang and Ion Stoica},
  booktitle={Proceedings of the ACM SIGOPS 29th Symposium on Operating Systems Principles},
  year={2023}
}

@misc{udrl0,
      title={Reinforcement Learning Upside Down: Don't Predict Rewards -- Just Map Them to Actions}, 
      author={Juergen Schmidhuber},
      year={2020},
      eprint={1912.02875},
      archivePrefix={arXiv},
      primaryClass={cs.AI},
      url={https://arxiv.org/abs/1912.02875}, 
}

@misc{udrl,
      title={Training Agents using Upside-Down Reinforcement Learning}, 
      author={Rupesh Kumar Srivastava and Pranav Shyam and Filipe Mutz and Wojciech Jaśkowski and Jürgen Schmidhuber},
      year={2021},
      eprint={1912.02877},
      archivePrefix={arXiv},
      primaryClass={cs.LG},
      url={https://arxiv.org/abs/1912.02877}, 
}

@misc{smollm3,
  title={{SmolLM3: smol, multilingual, long-context reasoner}},
  author={Bakouch, Elie and Ben Allal, Loubna and Lozhkov, Anton and Tazi, Nouamane and Tunstall, Lewis and Patiño, Carlos Miguel and Beeching, Edward and Roucher, Aymeric and Reedi, Aksel Joonas and Gallouédec, Quentin and Rasul, Kashif and Habib, Nathan and Fourrier, Clémentine and Kydlicek, Hynek and Penedo, Guilherme and Larcher, Hugo and Morlon, Mathieu and Srivastav, Vaibhav and Lochner, Joshua and Nguyen, Xuan-Son and Raffel, Colin and von Werra, Leandro and Wolf, Thomas},
  year={2025},
  howpublished={\url{https://huggingface.co/blog/smollm3}}
}

@misc{qwen3,
      title={Qwen3 Technical Report}, 
      author={An Yang and Anfeng Li and Baosong Yang and Beichen Zhang and Binyuan Hui and Bo Zheng and Bowen Yu and Chang Gao and Chengen Huang and Chenxu Lv and Chujie Zheng and Dayiheng Liu and Fan Zhou and Fei Huang and Feng Hu and Hao Ge and Haoran Wei and Huan Lin and Jialong Tang and Jian Yang and Jianhong Tu and Jianwei Zhang and Jianxin Yang and Jiaxi Yang and Jing Zhou and Jingren Zhou and Junyang Lin and Kai Dang and Keqin Bao and Kexin Yang and Le Yu and Lianghao Deng and Mei Li and Mingfeng Xue and Mingze Li and Pei Zhang and Peng Wang and Qin Zhu and Rui Men and Ruize Gao and Shixuan Liu and Shuang Luo and Tianhao Li and Tianyi Tang and Wenbiao Yin and Xingzhang Ren and Xinyu Wang and Xinyu Zhang and Xuancheng Ren and Yang Fan and Yang Su and Yichang Zhang and Yinger Zhang and Yu Wan and Yuqiong Liu and Zekun Wang and Zeyu Cui and Zhenru Zhang and Zhipeng Zhou and Zihan Qiu},
      year={2025},
      eprint={2505.09388},
      archivePrefix={arXiv},
      primaryClass={cs.CL},
      url={https://arxiv.org/abs/2505.09388}, 
}

@misc{gemma3,
      title={Gemma 3 Technical Report}, 
      author={Gemma Team and Aishwarya Kamath and Johan Ferret and Shreya Pathak and Nino Vieillard and Ramona Merhej and Sarah Perrin and Tatiana Matejovicova and Alexandre Ramé and Morgane Rivière and Louis Rouillard and Thomas Mesnard and Geoffrey Cideron and Jean-bastien Grill and Sabela Ramos and Edouard Yvinec and Michelle Casbon and Etienne Pot and Ivo Penchev and Gaël Liu and Francesco Visin and Kathleen Kenealy and Lucas Beyer and Xiaohai Zhai and Anton Tsitsulin and Robert Busa-Fekete and Alex Feng and Noveen Sachdeva and Benjamin Coleman and Yi Gao and Basil Mustafa and Iain Barr and Emilio Parisotto and David Tian and Matan Eyal and Colin Cherry and Jan-Thorsten Peter and Danila Sinopalnikov and Surya Bhupatiraju and Rishabh Agarwal and Mehran Kazemi and Dan Malkin and Ravin Kumar and David Vilar and Idan Brusilovsky and Jiaming Luo and Andreas Steiner and Abe Friesen and Abhanshu Sharma and Abheesht Sharma and Adi Mayrav Gilady and Adrian Goedeckemeyer and Alaa Saade and Alex Feng and Alexander Kolesnikov and Alexei Bendebury and Alvin Abdagic and Amit Vadi and András György and André Susano Pinto and Anil Das and Ankur Bapna and Antoine Miech and Antoine Yang and Antonia Paterson and Ashish Shenoy and Ayan Chakrabarti and Bilal Piot and Bo Wu and Bobak Shahriari and Bryce Petrini and Charlie Chen and Charline Le Lan and Christopher A. Choquette-Choo and CJ Carey and Cormac Brick and Daniel Deutsch and Danielle Eisenbud and Dee Cattle and Derek Cheng and Dimitris Paparas and Divyashree Shivakumar Sreepathihalli and Doug Reid and Dustin Tran and Dustin Zelle and Eric Noland and Erwin Huizenga and Eugene Kharitonov and Frederick Liu and Gagik Amirkhanyan and Glenn Cameron and Hadi Hashemi and Hanna Klimczak-Plucińska and Harman Singh and Harsh Mehta and Harshal Tushar Lehri and Hussein Hazimeh and Ian Ballantyne and Idan Szpektor and Ivan Nardini and Jean Pouget-Abadie and Jetha Chan and Joe Stanton and John Wieting and Jonathan Lai and Jordi Orbay and Joseph Fernandez and Josh Newlan and Ju-yeong Ji and Jyotinder Singh and Kat Black and Kathy Yu and Kevin Hui and Kiran Vodrahalli and Klaus Greff and Linhai Qiu and Marcella Valentine and Marina Coelho and Marvin Ritter and Matt Hoffman and Matthew Watson and Mayank Chaturvedi and Michael Moynihan and Min Ma and Nabila Babar and Natasha Noy and Nathan Byrd and Nick Roy and Nikola Momchev and Nilay Chauhan and Noveen Sachdeva and Oskar Bunyan and Pankil Botarda and Paul Caron and Paul Kishan Rubenstein and Phil Culliton and Philipp Schmid and Pier Giuseppe Sessa and Pingmei Xu and Piotr Stanczyk and Pouya Tafti and Rakesh Shivanna and Renjie Wu and Renke Pan and Reza Rokni and Rob Willoughby and Rohith Vallu and Ryan Mullins and Sammy Jerome and Sara Smoot and Sertan Girgin and Shariq Iqbal and Shashir Reddy and Shruti Sheth and Siim Põder and Sijal Bhatnagar and Sindhu Raghuram Panyam and Sivan Eiger and Susan Zhang and Tianqi Liu and Trevor Yacovone and Tyler Liechty and Uday Kalra and Utku Evci and Vedant Misra and Vincent Roseberry and Vlad Feinberg and Vlad Kolesnikov and Woohyun Han and Woosuk Kwon and Xi Chen and Yinlam Chow and Yuvein Zhu and Zichuan Wei and Zoltan Egyed and Victor Cotruta and Minh Giang and Phoebe Kirk and Anand Rao and Kat Black and Nabila Babar and Jessica Lo and Erica Moreira and Luiz Gustavo Martins and Omar Sanseviero and Lucas Gonzalez and Zach Gleicher and Tris Warkentin and Vahab Mirrokni and Evan Senter and Eli Collins and Joelle Barral and Zoubin Ghahramani and Raia Hadsell and Yossi Matias and D. Sculley and Slav Petrov and Noah Fiedel and Noam Shazeer and Oriol Vinyals and Jeff Dean and Demis Hassabis and Koray Kavukcuoglu and Clement Farabet and Elena Buchatskaya and Jean-Baptiste Alayrac and Rohan Anil and Dmitry and Lepikhin and Sebastian Borgeaud and Olivier Bachem and Armand Joulin and Alek Andreev and Cassidy Hardin and Robert Dadashi and Léonard Hussenot},
      year={2025},
      eprint={2503.19786},
      archivePrefix={arXiv},
      primaryClass={cs.CL},
      url={https://arxiv.org/abs/2503.19786}, 
}

@misc{phi4,
      title={Phi-4 Technical Report}, 
      author={Marah Abdin and Jyoti Aneja and Harkirat Behl and Sébastien Bubeck and Ronen Eldan and Suriya Gunasekar and Michael Harrison and Russell J. Hewett and Mojan Javaheripi and Piero Kauffmann and James R. Lee and Yin Tat Lee and Yuanzhi Li and Weishung Liu and Caio C. T. Mendes and Anh Nguyen and Eric Price and Gustavo de Rosa and Olli Saarikivi and Adil Salim and Shital Shah and Xin Wang and Rachel Ward and Yue Wu and Dingli Yu and Cyril Zhang and Yi Zhang},
      year={2024},
      eprint={2412.08905},
      archivePrefix={arXiv},
      primaryClass={cs.CL},
      url={https://arxiv.org/abs/2412.08905}, 
}

@misc{mistral,
      title={Mistral 7B}, 
      author={Albert Q. Jiang and Alexandre Sablayrolles and Arthur Mensch and Chris Bamford and Devendra Singh Chaplot and Diego de las Casas and Florian Bressand and Gianna Lengyel and Guillaume Lample and Lucile Saulnier and Lélio Renard Lavaud and Marie-Anne Lachaux and Pierre Stock and Teven Le Scao and Thibaut Lavril and Thomas Wang and Timothée Lacroix and William El Sayed},
      year={2023},
      eprint={2310.06825},
      archivePrefix={arXiv},
      primaryClass={cs.CL},
      url={https://arxiv.org/abs/2310.06825}, 
}

@misc{llama3,
      title={The Llama 3 Herd of Models}, 
      author={Abhimanyu Dubey and Abhinav Jauhri and Abhinav Pandey and Abhishek Kadian and Ahmad Al-Dahle and Aiesha Letman and Akhil Mathur and Alan Schelten and Amy Yang and Angela Fan and Anirudh Goyal and Anthony Hartshorn and Aobo Yang and Archi Mitra and Archie Sravankumar and Artem Korenev and Arthur Hinsvark and Arun Rao and Aston Zhang and Aurelien Rodriguez and Austen Gregerson and Ava Spataru and Baptiste Roziere and Bethany Biron and Binh Tang and Bobbie Chern and Charlotte Caucheteux and Chaya Nayak and Chloe Bi and Chris Marra and Chris McConnell and Christian Keller and Christophe Touret and Chunyang Wu and Corinne Wong and Cristian Canton Ferrer and Cyrus Nikolaidis and Damien Allonsius and Daniel Song and Danielle Pintz and Danny Livshits and David Esiobu and Dhruv Choudhary and Dhruv Mahajan and Diego Garcia-Olano and Diego Perino and Dieuwke Hupkes and Egor Lakomkin and Ehab AlBadawy and Elina Lobanova and Emily Dinan and Eric Michael Smith and Filip Radenovic and Frank Zhang and Gabriel Synnaeve and Gabrielle Lee and Georgia Lewis Anderson and Graeme Nail and Gregoire Mialon and Guan Pang and Guillem Cucurell and Hailey Nguyen and Hannah Korevaar and Hu Xu and Hugo Touvron and Iliyan Zarov and Imanol Arrieta Ibarra and Isabel Kloumann and Ishan Misra and Ivan Evtimov and Jade Copet and Jaewon Lee and Jan Geffert and Jana Vranes and Jason Park and Jay Mahadeokar and Jeet Shah and Jelmer van der Linde and Jennifer Billock and Jenny Hong and Jenya Lee and Jeremy Fu and Jianfeng Chi and Jianyu Huang and Jiawen Liu and Jie Wang and Jiecao Yu and Joanna Bitton and Joe Spisak and Jongsoo Park and Joseph Rocca and Joshua Johnstun and Joshua Saxe and Junteng Jia and Kalyan Vasuden Alwala and Kartikeya Upasani and Kate Plawiak and Ke Li and Kenneth Heafield and Kevin Stone and Khalid El-Arini and Krithika Iyer and Kshitiz Malik and Kuenley Chiu and Kunal Bhalla and Lauren Rantala-Yeary and Laurens van der Maaten and Lawrence Chen and Liang Tan and Liz Jenkins and Louis Martin and Lovish Madaan and Lubo Malo and Lukas Blecher and Lukas Landzaat and Luke de Oliveira and Madeline Muzzi and Mahesh Pasupuleti and Mannat Singh and Manohar Paluri and Marcin Kardas and Mathew Oldham and Mathieu Rita and Maya Pavlova and Melanie Kambadur and Mike Lewis and Min Si and Mitesh Kumar Singh and Mona Hassan and Naman Goyal and Narjes Torabi and Nikolay Bashlykov and Nikolay Bogoychev and Niladri Chatterji and Olivier Duchenne and Onur Çelebi and Patrick Alrassy and Pengchuan Zhang and Pengwei Li and Petar Vasic and Peter Weng and Prajjwal Bhargava and Pratik Dubal and Praveen Krishnan and Punit Singh Koura and Puxin Xu and Qing He and Qingxiao Dong and Ragavan Srinivasan and Raj Ganapathy and Ramon Calderer and Ricardo Silveira Cabral and Robert Stojnic and Roberta Raileanu and Rohit Girdhar and Rohit Patel and Romain Sauvestre and Ronnie Polidoro and Roshan Sumbaly and Ross Taylor and Ruan Silva and Rui Hou and Rui Wang and Saghar Hosseini and Sahana Chennabasappa and Sanjay Singh and Sean Bell and Seohyun Sonia Kim and Sergey Edunov and Shaoliang Nie and Sharan Narang and Sharath Raparthy and Sheng Shen and Shengye Wan and Shruti Bhosale and Shun Zhang and Simon Vandenhende and Soumya Batra and Spencer Whitman and Sten Sootla and Stephane Collot and Suchin Gururangan and Sydney Borodinsky and Tamar Herman and Tara Fowler and Tarek Sheasha and Thomas Georgiou and Thomas Scialom and Tobias Speckbacher and Todor Mihaylov and Tong Xiao and Ujjwal Karn and Vedanuj Goswami and Vibhor Gupta and Vignesh Ramanathan and Viktor Kerkez and Vincent Gonguet and Virginie Do and Vish Vogeti and Vladan Petrovic and Weiwei Chu and Wenhan Xiong and Wenyin Fu and Whitney Meers and Xavier Martinet and Xiaodong Wang and Xiaoqing Ellen Tan and Xinfeng Xie and Xuchao Jia and Xuewei Wang and Yaelle Goldschlag and Yashesh Gaur and Yasmine Babaei and Yi Wen and Yiwen Song and Yuchen Zhang and Yue Li and Yuning Mao and Zacharie Delpierre Coudert and Zheng Yan and Zhengxing Chen and Zoe Papakipos and Aaditya Singh and Aaron Grattafiori and Abha Jain and Adam Kelsey and Adam Shajnfeld and Adithya Gangidi and Adolfo Victoria and Ahuva Goldstand and Ajay Menon and Ajay Sharma and Alex Boesenberg and Alex Vaughan and Alexei Baevski and Allie Feinstein and Amanda Kallet and Amit Sangani and Anam Yunus and Andrei Lupu and Andres Alvarado and Andrew Caples and Andrew Gu and Andrew Ho and Andrew Poulton and Andrew Ryan and Ankit Ramchandani and Annie Franco and Aparajita Saraf and Arkabandhu Chowdhury and Ashley Gabriel and Ashwin Bharambe and Assaf Eisenman and Azadeh Yazdan and Beau James and Ben Maurer and Benjamin Leonhardi and Bernie Huang and Beth Loyd and Beto De Paola and Bhargavi Paranjape and Bing Liu and Bo Wu and Boyu Ni and Braden Hancock and Bram Wasti and Brandon Spence and Brani Stojkovic and Brian Gamido and Britt Montalvo and Carl Parker and Carly Burton and Catalina Mejia and Changhan Wang and Changkyu Kim and Chao Zhou and Chester Hu and Ching-Hsiang Chu and Chris Cai and Chris Tindal and Christoph Feichtenhofer and Damon Civin and Dana Beaty and Daniel Kreymer and Daniel Li and Danny Wyatt and David Adkins and David Xu and Davide Testuggine and Delia David and Devi Parikh and Diana Liskovich and Didem Foss and Dingkang Wang and Duc Le and Dustin Holland and Edward Dowling and Eissa Jamil and Elaine Montgomery and Eleonora Presani and Emily Hahn and Emily Wood and Erik Brinkman and Esteban Arcaute and Evan Dunbar and Evan Smothers and Fei Sun and Felix Kreuk and Feng Tian and Firat Ozgenel and Francesco Caggioni and Francisco Guzmán and Frank Kanayet and Frank Seide and Gabriela Medina Florez and Gabriella Schwarz and Gada Badeer and Georgia Swee and Gil Halpern and Govind Thattai and Grant Herman and Grigory Sizov and Guangyi and Zhang and Guna Lakshminarayanan and Hamid Shojanazeri and Han Zou and Hannah Wang and Hanwen Zha and Haroun Habeeb and Harrison Rudolph and Helen Suk and Henry Aspegren and Hunter Goldman and Ibrahim Damlaj and Igor Molybog and Igor Tufanov and Irina-Elena Veliche and Itai Gat and Jake Weissman and James Geboski and James Kohli and Japhet Asher and Jean-Baptiste Gaya and Jeff Marcus and Jeff Tang and Jennifer Chan and Jenny Zhen and Jeremy Reizenstein and Jeremy Teboul and Jessica Zhong and Jian Jin and Jingyi Yang and Joe Cummings and Jon Carvill and Jon Shepard and Jonathan McPhie and Jonathan Torres and Josh Ginsburg and Junjie Wang and Kai Wu and Kam Hou U and Karan Saxena and Karthik Prasad and Kartikay Khandelwal and Katayoun Zand and Kathy Matosich and Kaushik Veeraraghavan and Kelly Michelena and Keqian Li and Kun Huang and Kunal Chawla and Kushal Lakhotia and Kyle Huang and Lailin Chen and Lakshya Garg and Lavender A and Leandro Silva and Lee Bell and Lei Zhang and Liangpeng Guo and Licheng Yu and Liron Moshkovich and Luca Wehrstedt and Madian Khabsa and Manav Avalani and Manish Bhatt and Maria Tsimpoukelli and Martynas Mankus and Matan Hasson and Matthew Lennie and Matthias Reso and Maxim Groshev and Maxim Naumov and Maya Lathi and Meghan Keneally and Michael L. Seltzer and Michal Valko and Michelle Restrepo and Mihir Patel and Mik Vyatskov and Mikayel Samvelyan and Mike Clark and Mike Macey and Mike Wang and Miquel Jubert Hermoso and Mo Metanat and Mohammad Rastegari and Munish Bansal and Nandhini Santhanam and Natascha Parks and Natasha White and Navyata Bawa and Nayan Singhal and Nick Egebo and Nicolas Usunier and Nikolay Pavlovich Laptev and Ning Dong and Ning Zhang and Norman Cheng and Oleg Chernoguz and Olivia Hart and Omkar Salpekar and Ozlem Kalinli and Parkin Kent and Parth Parekh and Paul Saab and Pavan Balaji and Pedro Rittner and Philip Bontrager and Pierre Roux and Piotr Dollar and Polina Zvyagina and Prashant Ratanchandani and Pritish Yuvraj and Qian Liang and Rachad Alao and Rachel Rodriguez and Rafi Ayub and Raghotham Murthy and Raghu Nayani and Rahul Mitra and Raymond Li and Rebekkah Hogan and Robin Battey and Rocky Wang and Rohan Maheswari and Russ Howes and Ruty Rinott and Sai Jayesh Bondu and Samyak Datta and Sara Chugh and Sara Hunt and Sargun Dhillon and Sasha Sidorov and Satadru Pan and Saurabh Verma and Seiji Yamamoto and Sharadh Ramaswamy and Shaun Lindsay and Shaun Lindsay and Sheng Feng and Shenghao Lin and Shengxin Cindy Zha and Shiva Shankar and Shuqiang Zhang and Shuqiang Zhang and Sinong Wang and Sneha Agarwal and Soji Sajuyigbe and Soumith Chintala and Stephanie Max and Stephen Chen and Steve Kehoe and Steve Satterfield and Sudarshan Govindaprasad and Sumit Gupta and Sungmin Cho and Sunny Virk and Suraj Subramanian and Sy Choudhury and Sydney Goldman and Tal Remez and Tamar Glaser and Tamara Best and Thilo Kohler and Thomas Robinson and Tianhe Li and Tianjun Zhang and Tim Matthews and Timothy Chou and Tzook Shaked and Varun Vontimitta and Victoria Ajayi and Victoria Montanez and Vijai Mohan and Vinay Satish Kumar and Vishal Mangla and Vítor Albiero and Vlad Ionescu and Vlad Poenaru and Vlad Tiberiu Mihailescu and Vladimir Ivanov and Wei Li and Wenchen Wang and Wenwen Jiang and Wes Bouaziz and Will Constable and Xiaocheng Tang and Xiaofang Wang and Xiaojian Wu and Xiaolan Wang and Xide Xia and Xilun Wu and Xinbo Gao and Yanjun Chen and Ye Hu and Ye Jia and Ye Qi and Yenda Li and Yilin Zhang and Ying Zhang and Yossi Adi and Youngjin Nam and Yu and Wang and Yuchen Hao and Yundi Qian and Yuzi He and Zach Rait and Zachary DeVito and Zef Rosnbrick and Zhaoduo Wen and Zhenyu Yang and Zhiwei Zhao},
      year={2024},
      eprint={2407.21783},
      archivePrefix={arXiv},
      primaryClass={cs.AI},
      url={https://arxiv.org/abs/2407.21783}, 
}

@misc{gpt4,
      title={GPT-4 Technical Report}, 
      author={OpenAI and Josh Achiam and Steven Adler and Sandhini Agarwal and Lama Ahmad and Ilge Akkaya and Florencia Leoni Aleman and Diogo Almeida and Janko Altenschmidt and Sam Altman and Shyamal Anadkat and Red Avila and Igor Babuschkin and Suchir Balaji and Valerie Balcom and Paul Baltescu and Haiming Bao and Mohammad Bavarian and Jeff Belgum and Irwan Bello and Jake Berdine and Gabriel Bernadett-Shapiro and Christopher Berner and Lenny Bogdonoff and Oleg Boiko and Madelaine Boyd and Anna-Luisa Brakman and Greg Brockman and Tim Brooks and Miles Brundage and Kevin Button and Trevor Cai and Rosie Campbell and Andrew Cann and Brittany Carey and Chelsea Carlson and Rory Carmichael and Brooke Chan and Che Chang and Fotis Chantzis and Derek Chen and Sully Chen and Ruby Chen and Jason Chen and Mark Chen and Ben Chess and Chester Cho and Casey Chu and Hyung Won Chung and Dave Cummings and Jeremiah Currier and Yunxing Dai and Cory Decareaux and Thomas Degry and Noah Deutsch and Damien Deville and Arka Dhar and David Dohan and Steve Dowling and Sheila Dunning and Adrien Ecoffet and Atty Eleti and Tyna Eloundou and David Farhi and Liam Fedus and Niko Felix and Simón Posada Fishman and Juston Forte and Isabella Fulford and Leo Gao and Elie Georges and Christian Gibson and Vik Goel and Tarun Gogineni and Gabriel Goh and Rapha Gontijo-Lopes and Jonathan Gordon and Morgan Grafstein and Scott Gray and Ryan Greene and Joshua Gross and Shixiang Shane Gu and Yufei Guo and Chris Hallacy and Jesse Han and Jeff Harris and Yuchen He and Mike Heaton and Johannes Heidecke and Chris Hesse and Alan Hickey and Wade Hickey and Peter Hoeschele and Brandon Houghton and Kenny Hsu and Shengli Hu and Xin Hu and Joost Huizinga and Shantanu Jain and Shawn Jain and Joanne Jang and Angela Jiang and Roger Jiang and Haozhun Jin and Denny Jin and Shino Jomoto and Billie Jonn and Heewoo Jun and Tomer Kaftan and Łukasz Kaiser and Ali Kamali and Ingmar Kanitscheider and Nitish Shirish Keskar and Tabarak Khan and Logan Kilpatrick and Jong Wook Kim and Christina Kim and Yongjik Kim and Jan Hendrik Kirchner and Jamie Kiros and Matt Knight and Daniel Kokotajlo and Łukasz Kondraciuk and Andrew Kondrich and Aris Konstantinidis and Kyle Kosic and Gretchen Krueger and Vishal Kuo and Michael Lampe and Ikai Lan and Teddy Lee and Jan Leike and Jade Leung and Daniel Levy and Chak Ming Li and Rachel Lim and Molly Lin and Stephanie Lin and Mateusz Litwin and Theresa Lopez and Ryan Lowe and Patricia Lue and Anna Makanju and Kim Malfacini and Sam Manning and Todor Markov and Yaniv Markovski and Bianca Martin and Katie Mayer and Andrew Mayne and Bob McGrew and Scott Mayer McKinney and Christine McLeavey and Paul McMillan and Jake McNeil and David Medina and Aalok Mehta and Jacob Menick and Luke Metz and Andrey Mishchenko and Pamela Mishkin and Vinnie Monaco and Evan Morikawa and Daniel Mossing and Tong Mu and Mira Murati and Oleg Murk and David Mély and Ashvin Nair and Reiichiro Nakano and Rajeev Nayak and Arvind Neelakantan and Richard Ngo and Hyeonwoo Noh and Long Ouyang and Cullen O'Keefe and Jakub Pachocki and Alex Paino and Joe Palermo and Ashley Pantuliano and Giambattista Parascandolo and Joel Parish and Emy Parparita and Alex Passos and Mikhail Pavlov and Andrew Peng and Adam Perelman and Filipe de Avila Belbute Peres and Michael Petrov and Henrique Ponde de Oliveira Pinto and Michael and Pokorny and Michelle Pokrass and Vitchyr H. Pong and Tolly Powell and Alethea Power and Boris Power and Elizabeth Proehl and Raul Puri and Alec Radford and Jack Rae and Aditya Ramesh and Cameron Raymond and Francis Real and Kendra Rimbach and Carl Ross and Bob Rotsted and Henri Roussez and Nick Ryder and Mario Saltarelli and Ted Sanders and Shibani Santurkar and Girish Sastry and Heather Schmidt and David Schnurr and John Schulman and Daniel Selsam and Kyla Sheppard and Toki Sherbakov and Jessica Shieh and Sarah Shoker and Pranav Shyam and Szymon Sidor and Eric Sigler and Maddie Simens and Jordan Sitkin and Katarina Slama and Ian Sohl and Benjamin Sokolowsky and Yang Song and Natalie Staudacher and Felipe Petroski Such and Natalie Summers and Ilya Sutskever and Jie Tang and Nikolas Tezak and Madeleine B. Thompson and Phil Tillet and Amin Tootoonchian and Elizabeth Tseng and Preston Tuggle and Nick Turley and Jerry Tworek and Juan Felipe Cerón Uribe and Andrea Vallone and Arun Vijayvergiya and Chelsea Voss and Carroll Wainwright and Justin Jay Wang and Alvin Wang and Ben Wang and Jonathan Ward and Jason Wei and CJ Weinmann and Akila Welihinda and Peter Welinder and Jiayi Weng and Lilian Weng and Matt Wiethoff and Dave Willner and Clemens Winter and Samuel Wolrich and Hannah Wong and Lauren Workman and Sherwin Wu and Jeff Wu and Michael Wu and Kai Xiao and Tao Xu and Sarah Yoo and Kevin Yu and Qiming Yuan and Wojciech Zaremba and Rowan Zellers and Chong Zhang and Marvin Zhang and Shengjia Zhao and Tianhao Zheng and Juntang Zhuang and William Zhuk and Barret Zoph},
      year={2024},
      eprint={2303.08774},
      archivePrefix={arXiv},
      primaryClass={cs.CL},
      url={https://arxiv.org/abs/2303.08774}, 
}

\vspace{12pt}

\end{document}